\documentclass[runningheads]{llncs}
\usepackage{graphicx}
\usepackage{booktabs}
\usepackage{cite}
\usepackage{url}
\usepackage{comment}
\usepackage{multirow}
\usepackage[T2A]{fontenc}
\usepackage[utf8]{inputenc}
\usepackage[russian,english]{babel}
\usepackage{hyperref}

\begin{document}

\title{ELMo and BERT in semantic change detection for Russian}
\author{Julia Rodina\inst{1,*}\orcidID{0000-0002-0289-7777} \and
Yuliya Trofimova\inst{1,*}\orcidID{0000-0001-6163-4304} \and
Andrey Kutuzov\inst{2} \orcidID{0000-0003-2540-5912} \and
Ekaterina Artemova \inst{1}\orcidID{0000-0003-4920-1623} }

\authorrunning{Julia Rodina, Yuliya Trofimova, et al.}
\institute{National Research University Higher School of Economics, Moscow, Russia \and  
University of Oslo, Oslo, Norway \\
\email{julia.rodina97@gmail.com}, \email{djuliya.trofimova@gmail.com}, \email{andreku@ifi.uio.no}, \email{echernyak@hse.ru}\\ 
$*$ These authors contributed equally to this work.}
\maketitle
\begin{abstract}
We study the effectiveness of contextualized embeddings for the task of diachronic semantic change detection for Russian language data. Evaluation test sets consist of Russian nouns and adjectives annotated based on their occurrences in texts created in pre-Soviet, Soviet and post-Soviet time periods. ELMo and BERT architectures are compared on the task of ranking Russian words according to the degree of their semantic change over time. We use several methods for aggregation of contextualized embeddings from these architectures and evaluate their performance. Finally, we compare unsupervised and supervised techniques in this task.
\end{abstract}

\section{Introduction}

In this research, we apply ELMo (Embeddings from Language Models) \cite{elmo} and BERT (Bidirectional Encoder Representations from Transformers) \cite{bert} models fine-tuned on historical corpora of Russian language to estimate diachronic semantic changes. We do that by extracting contextualized word representations for each time bin and quantifying their differences. This approach is data-driven and does not require any manual intervention.
For evaluation, we use human-annotated datasets and show that contextualized embedding models can be used to rank words by the degree of their semantic change, yielding a significant correlation with human judgments. 

It is important to mention that we treat the word meaning as a function of the word's contexts in natural language texts. This concept corresponds to the distributional hypothesis \cite{firth}.

The rest of the paper is organized as follows: in Section~\ref{sec:background}, we describe previous research on the automatic lexical semantic change detection. In Section~\ref{sec:data}, we present natural language data used in our research, and the dataset structure. The training of contextualized embeddings and models' architecture is described in Section~\ref{sec:experiments}. In this section, we also describe and apply various algorithms for semantic change detection. In Section~\ref{sec:results}, we calculate correlation of their results with human judgments to evaluate their performance. In Section~\ref{sec:conclusion}, we summarize our contributions and discuss possibilities for future research.

\section{Related work}
\label{sec:background}

The nature and reasons of semantic change processes have been studied in linguistics for a long time, at least since the theoretical work of \cite{breal} where the cognitive laws of semantic change were formulated. Later there were other works on categorizing different types of semantic changes \cite{bloomfield, stern}. 

With the increasing amount of available language data, researchers started to focus on empirical approaches to semantic change in addition to theoretical work. Earlier studies consisted primarily of corpus-based analysis (\cite{michel, duh} among many others), and used raw word frequencies to detect semantic shifts. However, there already were applications of distributional methods, for example, Temporal Random Indexing \cite{ri}, co-occurrences matrices weighted by Local Mutual Information \cite{gulordava}, graph-based methods \cite{mitra} and others. 

Among the works focusing on Russian, one can mention a recent book \cite{twentywords} in which 20 words were manually analyzed by exploring the contexts in which they appear and counting the number of their uses for each period. This allowed to picture the history of changes of meanings which words undergo.

\subsection{Static word embeddings}
After the widespread usage of word embeddings \cite{mikolov_2} had started, the focus has shifted to detecting semantic changes using dense distributional word representations; see \cite{kim, kulkarni, hamilton-law, rosenfeld} and many others.
Word embeddings represent meaning of words as dense vectors learned from their co-occurrences counts in the training corpora. This approach has gained popularity as it can leverage un-annotated natural language data and produces both efficient and easy to work with continuous representations of meaning. 
Comprehensive surveys on research about semantic change detection using ‘static’ word embeddings are given in \cite{kutuzov-survey} and \cite{tang}.

\cite{twentywords} inspired the first (to our knowledge) publication which applied static word embedding models to Russian data in order to detect diachronic lexical changes \cite{kutuzov-kuzmenko}. The authors compared sets of word’s nearest neighbors and concluded that Kendall’s $\tau$ and Jaccard distance worked best in scoring changes. More recent work \cite{fomin} extended this research to more granular time bins (periods of 1 year): they analyzed semantic shifts between static embeddings trained on yearly corpora of Russian news texts from the year 2000 up to 2014. They evaluated 5 algorithms for semantic shift detection and provided solid baselines for future research in Russian language. Note, however, that \cite{fomin} solved the classification task (whether a word has changed its meaning or not), while in the present paper we solve the ranking task (which words have changed more or less than the others). Also, the test set we use (\textit{RuSemShift}) is much more extensive and consistent (see subsection~\ref{rusemshift}).

\subsection{Contextualized word embeddings}
Static word embeddings assign the same vector representation to each word (lemma) occurrence: they produce context-independent vectors at the inference time. However, recent advances in natural language processing made it possible to implement models that allow to obtain higher quality \textit{contextualized} representations. 
The main difference of contextualized models is that at the inference stage tokens are assigned different embeddings depending on their context in the input data. This results in richer word features and more realistic word representations. There are several recently published contextualized architectures based on language modeling task (predicting the next most probable word given a sequence of tokens) \cite{ulmfit, elmo, gpt, bert}.

This provides new opportunities for diachronic analysis: for example, it is possible to group similar token representations and measure a diversity of such representations, while predefined number of senses is not strictly necessary. Thus, currently there is an increased interest in the topic of language change detection using contextualized word embeddings \cite{hu, mario-master, mario, martinc, martinc-clusters, prt}.

\cite{hu} used a list of polysemous words with predefined set of senses, then a pre-trained BERT model was applied to diachronic corpora to extract token embeddings that are the closest to the predefined sense embedding. Evolution of each word was measured by comparing distributions of senses in different time slices. In \cite{mario-master} and \cite{mario}, a pre-trained BERT model was used to obtain representations of word usages in an unsupervised fashion, without predefined list or number of senses. Representations with similar usages then were clustered using k-Means algorithm and distributions of word’s usages in these clusters were used in two metrics for quantifying the degree of semantic change: entropy difference and Jensen-Shannon divergence. They also used average pairwise distance, that does not need clusterization and only requires usage matrices from two time periods. 

\cite{martinc} used averaged time-specific BERT representations and calculated cosine distance between averaged vectors of two time periods as a measure of semantic change. \cite{martinc-clusters} tested Affinity Propagation algorithm for usage clusterization and showed that it is consistently better than k-Means. Finally, \cite{prt} applied approaches similar to \cite{mario}, but also analyzing ELMo models and adding cosine similarity of average vectors as a measure. Their algorithms were evaluated on Subtask 2 (ranking) of SemEval-2020 Task 1 \cite{semeval} for four different languages and strongly outperformed the baselines. We will test each of these clustering methods and  cosine similarity of averaged vectors for our task in the present paper.

We decided to compare BERT against ELMo. Most of the above mentioned works used pre-trained BERT language models, but ELMo allows faster training and inference which makes it easier to train diachronic models completely on one's own data.

\section{Data}
\label{sec:data}

\subsection{Corpora}

For both BERT and ELMo architectures, we used pre-trained models from prior work.
The \texttt{RuBERT} model is a Multilingual BERT \cite{bert} fine-tuned on the Russian Wikipedia and news articles (about 850 million tokens) \cite{rubert}. The \texttt{tayga\_lemmas\_elmo\_2048\_2019} is an ELMo model trained on the Taiga corpus of Russian (almost 5 billion tokens) \cite{taiga}; it is available from the RusVectores web service \cite{rusvectores}.

Both models were additionally fine-tuned on the Russian National Corpus (RNC) which contains texts in Russian language produced from the middle of the XVIII to the beginning of the XXI century (about 320 million word tokens in total, including punctuation). 
Before the fine-tuning, the corpus was segmented into sentences, then tokenized and lemmatized with the Universal Dependencies model trained on the SynTagRus Russian UD treebank \cite{ud_rus} using UDPipe 1.2 \cite{udpipe}.
This was motivated by recent research \cite{kutuzov-kuzmenko-lemma} which showed that for Russian data, lemmatization improves the results of contextualized architectures on the word sense disambiguation task. Reducing the noise effect of word forms is useful for tracing semantic shifts among all occurrences of the word in the aggregate.

For the purposes of diachronic semantic change detection, the RNC corpus was divided into three parts, with the boundaries between their consequent pairs corresponding to the rise and fall of the Soviet Union, which undoubtedly brought along major social, cultural and political shifts:
\begin{enumerate}
    \item texts produced before 1917 (pre-Soviet period): 94 million tokens;
    \item texts produced in 1918-1990 (Soviet period): 123 million tokens;
    \item texts produced in 1991-2017 (post-Soviet period): 107 million tokens.
\end{enumerate}
For extracting time-specific word's embeddings at the inference stage, we used each time-specific sub-corpus separately.

\subsection{Evaluation dataset} \label{rusemshift}

We used two human-annotated datasets of Russian nouns and adjectives for evaluation of diachronic semantic change detection methods. They are part of a larger \textit{RuSemShift} dataset\footnote{\url{https://github.com/juliarodina/RuSemShift}} and cover both consequent pairs of the RNC sub-corpora: from the pre-Soviet through the Soviet times ($RuSemShift_1$) and from the Soviet through the post-Soviet time ($RuSemShift_2$).

The datasets contain words annotated similar to the DURel framework \cite{durel} according to the intensity of the semantic changes they have undergone. Each word was presented to 5 independent annotators along with two context sentences. These pairs of sentences were sampled from the RNC and divided into three equal groups: EARLIER (both sentences are from the earlier time period), LATER (both sentences from the later time period) and COMPARE (one sentence from the earlier period and another from the later period). The annotators were asked to rate how different are the word's meanings in two sentences on the scale from 1 (unrelated meanings) to 4 (identical meanings). 

There are two measures for quantifying diachronic semantic change based on this manual annotation: 
\begin{enumerate}
    \item $\Delta$LATER which is the subtraction of the EARLIER group’s mean difference value from the LATER group’s mean difference value: 
    
    $\Delta$LATER(\textit{w}) $= Mean_{later}(w) - Mean_{earlier}(w)$;
    \item COMPARE which is the word’s mean difference value in the COMPARE group: 
    
    COMPARE(\textit{w}) = $Mean_{compare}$(\textit{w}).
\end{enumerate}

Note that we use two test sets, and thus two pairs of time periods. In each pair, the EARLIER and LATER groups stand for different time periods. Foe example, in $RuSemShift_2$, the EARLIER group corresponds to texts from the Soviet times, and the LATER group corresponds to texts from the post-Soviet times.

The experiments were performed on the filtered versions of the datasets where the words with the annotators' agreement lower than 0.2 were excluded. Their sizes are 48 words ($RuSemShift_1$) and 51 words ($RuSemShift_2$).

\section{Experimental setup}
\label{sec:experiments}

\subsection{Contextualized Word Embedding Models}

\subsubsection{ELMo}
is a deep character based bidirectional language model (biLM): a combination of two-layer long short-term memory (LSTM) networks on top of a convolutional layer with max-pooling \cite{elmo}. Having a pre-trained biLM model, we can obtain word representations that are learned functions of the internal state of this model and contain information about word’s syntax and semantics as well as its polysemy. The model we used (\texttt{tayga\_lemmas\_elmo\_2048\_2019}) was trained on the Taiga corpus for 3 epochs, with batch size 192. All the ELMo hyperparameters were left for their default values, except for the number of negative samples per batch which was reduced to 4 096 from 8 192 used in \cite{elmo}. After fine-tuning on the whole RNC, for each word from the test sets we extracted its contextualized token representations for every occurrence. We used only the top LSTM layer, since higher layer representations are more oriented to learn semantics and capture longer range dependencies \cite{elmo2}. 

\subsubsection{BERT}
(Bidirectional Encoder Representations from Transformers) makes use of Transformer \cite{DBLP:journals/corr/VaswaniSPUJGKP17}, an attention mechanism that learns contextual relations between words in a sentence. Masked language modeling technique allows bidirectional training of Transformer that allegedly allows for a deeper sense of language context. Training BERT from scratch is a computationally expensive process and demands huge resources. Fine-tuning is a well-known technique for adapting pre-trained BERT models to a specific corpus or task. We fine-tuned the \texttt{RuBERT} model on the RNC, before produce contextualized word representations. The model was being fine-tuned for 3 epochs, with the batch size 16. The original \texttt{RuBERT} has 12 heads and its hidden size is 768.
Then, for each word from our test sets, we extracted contextualized embedding for each of its occurrences during the respective time periods. We used the last BERT layer representations as token embeddings. 

For both ELMo and BERT, this resulted in three matrices of token embeddings, corresponding to three time periods. 

\subsection{Methods}

To evaluate ELMo and BERT in semantic change detection task, we extracted word embeddings for each word's usage from the RNC sub-corpora corresponding to the time periods under analysis. Recall that token embeddings are context-dependent. We applied four different aggregation methods to estimate semantic change degrees based on the token embeddings. Then we calculated correlation between their outputs and two gold human-annotated measures from the $RuSemShift_1$ and $RuSemShift_2$: $\Delta$LATER and COMPARE. We discuss the aggregation methods below.

\subsubsection{Inverted cosine similarity over word prototypes (PRT)}
The PRT method outperformed others in \cite{prt} and is computationally inexpensive.
First, we average all $n$ token embeddings of a given word in a specific sub-corpus, this producing a `prototypical' representation of the word. Then we calculate cosine distance between the average embeddings $u$ and $v$ from two different sub-corpora as a measure of semantic change:
\begin{equation}
\mathop{PRT} = 1 - cos(u, v) = 1 - \frac{\sum_{i=1}^n u_i \cdot v_i}{\sqrt{\sum_{i=1}^n u_i^2} \cdot \sqrt{\sum_{i=1}^n v_i^2}}
\label{eq:prt}
\end{equation}

\subsubsection{Clustering}
As mentioned above, ELMo and BERT return context-dependent token embeddings, and we need to compare them in order to estimate semantic shift. Another way to do it is clustering token embeddings. We can move away from comparing individual vectors or their averages, and instead look at word occurrences as belonging to several clusters. It is assumed that these clusters correspond to word senses. Following the previous work, we tested two clustering methods: Affinity Propagation \cite{aff} and k-Means \cite{journals/tit/Lloyd82}. One needs to manually set the hyperparameter for the number of clusters for k-Means (we empirically found $5$ to be the best value for our data). At the same time, Affinity Propagation clustering doesn't require a predetermined number of clusters, it is inferred automatically from data. 

We need to compare the resulting clusters of word's usages in two time periods. We used the method proposed in \cite{mario-master}. Namely, we clustered all embeddings of each word $w$  from the two time periods $(u_1^t, u_2^t, ..., u_n^t)$ and $(u_1^{t + 1}, u_2^{t + 1}, ..., u_m^{t + 1})$. Then, for each of the time bins $t$ and $t + 1$ we calculated the following distribution \ref{eq:like} which indicates the likelihood of encountering the word in a specific sense:
\begin{equation}
p_w^t[k] = \frac{|y_w^t[i] \in y_w^t,~ if~ y_w^t[i] = k|}{N_w^t}, ~~~y_w^t \in [1, K_w]^{N^t_w} 
\label{eq:like}
\end{equation}

where $y_w^t$ are the labels after clustering, $K$ is the number of clusters, and $N_w^t$ is the number of $w$ occurrences in the time period $t$.
These distributions are comparable, and we test two different methods for this:

\begin{enumerate}
    \item \textbf{Jensen-Shannon divergence (JSD)}. Here, we compute the JSD between two distributions: 
    \begin{equation}
    \mathop{JSD} = \sqrt{0.5 \cdot (\mathop{KL}(p||m) + KL(q||m))}
    \label{eq:jsd}
    \end{equation}
    where $KL$ is Kullback-Leibler divergence \cite{Kullback51klDivergence}, $p$ and $q$ are sense distributions and $m$ is the point-wise mean of $p$ and $q$. 
    Higher JSD score indicates more intense change in the proportions of clustered word usage types across time periods.
    \item \textbf{Maximum square}. Here, we assume that slight changes in sense distribution may occur due to noise and do not manifest a real semantic change. At the same time, strong changes in context distribution may indicate serious semantic shifts. Therefore we apply the hand-picked function \ref{eq:ms}:
    \begin{equation}
    \mathop{MS} = max(\mathop{square}(p - q))
    \label{eq:ms}
    \end{equation}
\end{enumerate}

\section{Results}
\label{sec:results}

As mentioned above, we extracted all word usages from the RNC diachronic sub-corpora for each word from the filtered \textit{RuSemShift} datasets. Then we produced their contextualized representations and applied the aggregation methods described in Section~\ref{sec:experiments}. Due to the increasing computational complexity, we have limited the number of usages to 10 000 (by random sampling) for the clustering methods. 

Tables~\ref{tab:scores1} and \ref{tab:scores2} show the Spearman correlation coefficients between the models' and annotators' rankings for $RuSemShift_1$ and $RuSemShift_2$ test sets correspondingly. `MS' stands for the Maximum Square method. Asterisk indicates statistically significant correlations (p-value $>$ 0.05), bold highlighting indicates best scores for each measure. COMPARE scores in the tables are actually equal to $1-\mathop{COMPARE}$, because of the nature of this measure: lower COMPARE score means stronger change.  

For both datasets, the best aggregation method was the PRT, but clustering was not much inferior.  As for the functions for comparing distributions of tokens in clusters, the Jensen-Shannon divergence showed stronger correlation with human judgments than the hand-picked function of maximum square.

If we compare BERT and ELMo results, we can see that they are somewhat similar and it cannot be concluded that one model is better than another (despite ELMo having much less parameters). Also, one can notice from the tables that the models' rankings in general are more correlated with the ranking by the COMPARE measure than by the $\Delta$LATER measure. Thus, the COMPARE measure is easier to approximate. Statistically significant correlations for $RuSemShift_2$ are generally higher that for $RuSemShift_1$. It can be due to lexical specificity of the datasets.

An interesting example of semantic change is the word '\foreignlanguage{russian}{провальный}' ('\textit{failed}'). According to the human annotation, this word has a very strong degree of change when comparing the Soviet period to the post-Soviet period. BERT and ELMo with PRT also place it at the top positions among the words ranked by their semantic change degrees. Indeed, in the Soviet times, '\foreignlanguage{russian}{провальный}' was mostly used in the literal
meaning of '\foreignlanguage{russian}{провал}' – '\textit{a place where the surface collapsed inward}' or figurative meaning '\textit{loss of consciousness}' especially in the collocation '\foreignlanguage{russian}{провальный сон}' ('\textit{deep dream}'). In the Soviet period, its primary sense shifted to the more common nowadays meaning ('\textit{failed}'). 

In some cases, the models estimates do not correlate with human judgments. For example, most of our approaches yield high semantic change degree for the word `\foreignlanguage{russian}{дождь}' (`\textit{rain}') when comparing the pre-Soviet to the Soviet period. However, human annotation positions it quite low in the semantic change rankings for this period pair. Whether this is an error of the model or an insufficiency of the dataset, remains yet to be solved.

\begin{table}
\begin{minipage}{0.5\textwidth}
\begin{tabular}{l|l|ll|}
\toprule
\multicolumn{2}{c}{\textbf{Algorithms}} & \multicolumn{2}{c|}{\textbf{Spearman $\rho$}} \\
\midrule
 & \textbf{Measure} & ELMo & BERT \\ 
\midrule
\multicolumn{1}{c|}{\multirow{2}{*}{PRT}} & $\Delta$LATER & 0.200 & 0.346* \\  
\multicolumn{1}{c|}{} & \multicolumn{1}{c|}{COMPARE} & 0.409* & \textbf{0.490*} \\
\midrule
\multicolumn{1}{c|}{\multirow{2}{*}{Affinity/JSD}} & $\Delta$LATER & \textbf{0.406*} & 0.160 \\ 
\multicolumn{1}{c|}{} & \multicolumn{1}{c|}{COMPARE} & 0.276 & 0.295* \\ 
\midrule
\multicolumn{1}{c|}{\multirow{2}{*}{kMeans/JSD}} & $\Delta$LATER & 0.250 & 0.270 \\  
\multicolumn{1}{c|}{} & \multicolumn{1}{c|}{COMPARE} & 0.340* & 0.440* \\
\midrule
\multicolumn{1}{c|}{\multirow{2}{*}{kMeans/MS}} & $\Delta$LATER & 0.060 & 0.240 \\  
\multicolumn{1}{c|}{} & \multicolumn{1}{c|}{COMPARE} & 0.380* & 0.358* \\ 
\bottomrule
\end{tabular}
\caption{Correlations of models' predictions with {$RuSemShift_1$} annotations (change between the pre-Soviet and Soviet time periods).}     
\label{tab:scores1}
\end{minipage}
\begin{minipage}{0.5\textwidth}
\begin{tabular}{l|l|ll}
\toprule
\multicolumn{2}{c}{\textbf{Algorithms}} & \multicolumn{2}{c}{\textbf{Spearman $\rho$}} \\
 \midrule
 & \textbf{Measure} & ELMo & BERT \\ 
\midrule
\multicolumn{1}{c|}{\multirow{2}{*}{PRT}} & $\Delta$LATER & \textbf{0.300*} & 0.230 \\
\multicolumn{1}{c|}{} & \multicolumn{1}{c|}{COMPARE} & \textbf{0.557*} & 0.500* \\
\midrule
\multicolumn{1}{c|}{\multirow{2}{*}{Affinity/JSD}} & $\Delta$LATER & 0.200 & 0.136 \\  
\multicolumn{1}{c|}{} & \multicolumn{1}{c|}{COMPARE} & 0.363* & 0.408* \\ 
\midrule
\multicolumn{1}{c|}{\multirow{2}{*}{kMeans/JSD}} & $\Delta$LATER & 0.200 & 0.130 \\ 
\multicolumn{1}{c|}{} & \multicolumn{1}{c|}{COMPARE} & 0.535* & 0.480* \\ 
\midrule
\multicolumn{1}{c|}{\multirow{2}{*}{kMeans/MS}} & $\Delta$LATER & 0.074 & 0.120 \\ 
\multicolumn{1}{c|}{} & \multicolumn{1}{c|}{COMPARE} & 0.436* & 0.420* \\
\bottomrule
\end{tabular}
\caption{Correlations of models' predictions with $RuSemShift_2$ annotations (change between the Soviet and post-Soviet time periods).}     
\label{tab:scores2}
\end{minipage}
\end{table}

\section{Conclusion}
\label{sec:conclusion}

In this work, we evaluated how semantic change detection methods based on contextualized word representations from BERT and ELMo perform for Russian language data (diachronic sub-corpora of the Russian National Corpus). In particular, we tested them in the task of automatically ranking Russian words according to the manually annotated degree of their diachronic semantic change. 

Pre-trained ELMo and BERT models were fine-tuned on the full RNC corpus to make comparison as fair as possible. Then we applied several algorithms for semantic shift detection: cosine similarity on a word prototypes (PRT) and clustering algorithms together with measures for comparing word's usages distributions. 

For the second method, we applied two clustering algorithms: Affinity Propagation and k-Means. K-Means turned out to be generally better than Affinity Propagation.   However, the PRT method, using simple cosine similarity between averaged token embeddings (word prototypes) outperformed clustering algorithms in most cases and therefore suits better for this task.

To sum up, we showed that contextualized word representation models have significant correlation with human judgments in diachronic semantic change detection for Russian. Also we found out that there is not much difference between BERT and ELMo contextualized embeddings in this respect and we can't say that one architecture is significantly  better than another.

In the future work, it would be interesting ti make a more fair comparison between ELMo and BERT on the task of semantic change detection by pre-training both models on identical corpora from scratch. However, this will require significant computational resources in the case of BERT.

\section*{Acknowledgments}
This research was supported by the Russian Science Foundation grant 20-18-00206.

\bibliographystyle{splncs04}
\bibliography{paper}

\end{document}